\documentclass[11pt,a4paper]{article}
\usepackage[hyperref]{acl2020}
\usepackage{times}

\usepackage{graphicx}
\usepackage{arydshln}
\usepackage{booktabs}

\usepackage{url}
\newcommand{\ignore}[1]{}
\newcommand{\miguelcomment}[1]{\textcolor{blue}{\textbf{[#1 --\textsc{miguel}]}}}

\newcommand{\yogarshicomment}[1]{\textcolor{green}{\textbf{[#1 --\textsc{yogarshi}]}}}
\newcommand{\nimacomment}[1]{\textcolor{cyan}{\textbf{[#1 --\textsc{nima}]}}}
\newcommand{\shuaicomment}[1]{\textcolor{purple}{\textbf{[#1 --\textsc{shuai}]}}}
\usepackage{lipsum}

\newcommand\blfootnote[1]{%
  \begingroup
  \renewcommand\thefootnote{}\footnote{#1}%
  \addtocounter{footnote}{-1}%
  \endgroup
}

\aclfinalcopy 

\title{Severing the Edge Between Before and After: \\ Neural Architectures for Temporal Ordering of Events} 

\author{Miguel Ballesteros ~~~~ Rishita Anubhai ~~~~ Shuai Wang ~~~~ \\ \textbf{Nima Pourdamghani}  ~~~~ \textbf{Yogarshi Vyas}  ~~~~ \textbf{Jie Ma}~~~~ \\ \textbf{Parminder Bhatia} ~~~~ \textbf{Kathleen McKeown}$^*$~~~~\textbf{Yaser Al-Onaizan} \\
  Amazon AI \\
  {\tt \footnotesize{\{ballemig,ranubhai,wshui,nimpourd,yogarshi,jieman,parmib,mckeownk,onaizan\}@amazon.com}}}

\date{}

\begin{document}
\maketitle
\begin{abstract}
In this paper, we propose a neural architecture and a set of training methods for ordering events by predicting temporal relations. Our proposed models receive a pair of events within a span of text as input and they identify temporal relations (\textit{Before}, \textit{After}, \textit{Equal}, \textit{Vague}) between them. Given that a key challenge with this task is the scarcity of annotated data, our models rely on either pretrained representations (i.e. RoBERTa, BERT or ELMo), transfer and multi-task  learning (by leveraging complementary datasets), and self-training techniques. Experiments on the MATRES dataset of English documents establish a new state-of-the-art on this task. \blfootnote{$^*$Kathleen McKeown is an Amazon Scholar and a Professor at Columbia University.} 
\end{abstract}

\section{Introduction}

The task of temporal ordering of events involves predicting the temporal relation between a pair of input events in a span of text (Figure \ref{example}). This task is challenging as it requires deep understanding of temporal aspects of language and the amount of annotated data is scarce. 

\ignore{The task of temporal ordering of events involves predicting the temporal relation between a pair of input events (Figure \ref{example}). This task is challenging as it requires deep understanding of language and the amount of annotated data for the task is scarce. }


\begin{figure}[!ht]
\fbox{
\begin{minipage}{7.4cm}
Albright \textbf{(e1, came)} to the State Department to \textbf{(e2, offer)} condolences.
\end{minipage}
}
\caption{Example from the MATRES dataset. The relation between \textbf{(e1, came)} and \textbf{(e2, offer)} is \textit{Before}. Note that for the same span there may be other relation pairs.}
\label{example}
\end{figure}

The MATRES dataset \cite{ning-etal-2018-multi}  has become a de facto standard for temporal ordering of events.\footnote{\url{https://github.com/qiangning/MATRES}} It contains 13,577 pairs of events annotated with a temporal relation (\textit{Before}, \textit{After}, \textit{Equal}, \textit{Vague}) within 256 \textbf{English} documents (and 20 more for evaluation) from TimeBank\footnote{\url{https://catalog.ldc.upenn.edu/LDC2006T08}} \cite{pustejovsky2003timebank}, AQUAINT\footnote{\url{https://catalog.ldc.upenn.edu/LDC2002T31}} \cite{graffaquaint} and
Platinum \cite{uzzaman-etal-2013-semeval}. \ignore{\yogarshicomment{I think this paragraph is too much detail too soon and delays getting to our contributions. I would keep only the salient properties of MATRES here (or really, only the example), and move everything else later on in either MTL section or experiments section}} 

In this paper, we present a set of neural architectures for temporal ordering of events. Our main model
(Section \ref{ourmodel}) is similar to the temporal ordering models designed by \newcite{goyal-durrett-2019-embedding}, \newcite{liu-etal-2019-attention} and \newcite{ning-etal-2019-improved}.  

Our main contributions are: (1) a neural architecture that can flexibly adapt different encoders and pretrained word embedders to form a contextual pairwise argument representation. Given the scarcity of training data, (2) we explore the application of an existing framework for Scheduled Multitask-Learning (henceforth SMTL) \cite{kiperwasser-ballesteros-2018-scheduled} by leveraging complementary (temporal and non temporal) information to our models; this imitates pretraining and finetuning. This consumes timex information in a different way than \newcite{goyal-durrett-2019-embedding}. (3) A self-training method that incorporates the predictions of our model and learns from them; we test it jointly with the SMTL method.
 
Our baseline model that uses RoBERTa \cite{liu2019roberta} already surpasses the state-of-the-art by 2 F1 points. Applying SMTL techniques affords further improvements with at least one of our auxiliary tasks. Finally, our self-training experiments, explored via SMTL as well, establishes yet another state-of-the-art yielding a total improvement of almost 4 F1 points over results from past work.

\section{Our Baseline Model}
\label{ourmodel}

Our pairwise temporal ordering model receives as input a sequence $X_{[0,n)}$ of $n$ tokens (or subword units for BERT-like models) i.e. $\{x_0, x_1, ..., x_{n-1}\}$,  representing the input text. A subsequence \textit{span}$_i$ is defined by \textit{start}$_i$, \textit{end}$_i \in \mathopen[0, n\mathclose)$. Subsequences \textit{span}$_1$ and \textit{span}$_2$ represent the input pair of argument events $e1$ and $e2$ respectively. The goal of the model is to predict the temporal relation between $e1$ and $e2$.

First, the model embeds the input sequence into a vector representation using either static wang2vec representations \cite{ling-etal-2015-two}, or contextualized representations from ELMo \cite{peters-etal-2018-deep}, BERT \cite{devlin-etal-2019-bert}, or RoBERTa \cite{liu2019roberta}. These embedded sequences are then optionally encoded with either LSTMs or Transformers. When BERT or RoBERTa is used to embed the input, we do not use any sequence encoders\ignore{(with the exception of ELMo, for which we tried both LSTMs and Transformers)}. The final sequence representation $H_{[0,n)}$ comprises of individual token representations i.e. $\{h_0, h_1, ..., h_{n-1}\}$.

While the goal is to predict the temporal relation between \textit{span}$_1$ and \textit{span}$_2$, the context around these two spans also has linguistic signals that connect the two arguments. To use this contextual information, we extract five constituent subsequences from the sequence representation $H_{[0,n)}$: (1) $S_1$, the subsequence before \textit{span}$_1$ i.e., $H_{[0 , \textit{start}_1)}$, (2) $S_2$, the subsequence corresponding to \textit{span}$_1$ i.e., $H_{[\textit{start}_1 , \textit{end}_1)}$, (3) $S_3$, the subsequence between \textit{span}$_1$ and \textit{span}$_2$ i.e, $H_{[\textit{end}_1 , \textit{start}_2)}$, 
(4) $S_4$, the subsequence corresponding to \textit{span}$_2$ i.e., $H_{[\textit{start}_2,\textit{end}_2)}$ and (5) $S_5$, the subsequence after \textit{span}$_2$, i.e. $H_{[\textit{end}_2, n)}$. Each of these subsequences $S_i$ has a variable number of tokens which are pooled to yield a fixed size representation $s_i$:
\begin{equation}\label{eq:2}
s_i = \textit{pool}(S_i) \ \ \forall i \in \{1, ... , 5\}
\end{equation}
where \textit{pool} is the result of concatenating the output of an attention mechanism (we use the \textit{word attention} pooling method \cite{yang-etal-2016-hierarchical} for all tokens in a given span) and mean pooling.

The final contextual pair representation \textit{c} is formed by concatenating\footnote{$\odot$ is used to denote concatenation} the five span representations $s_i$ with a sequence representation $r$. For models with BERT and RoBERTa, $r$ is the CLS and $<$s$>$ token representation respectively while for other models $r =$ \textit{pool}($H_{[0,n)}$). 
\begin{equation}\label{eq:2}
c = \textit{s}_1 \odot \textit{s}_2 \odot \textit{s}_3 \odot \textit{s}_4 \odot \textit{s}_5 \odot \textit{r}
\end{equation}
This final contextual pair representation \textit{c} is then projected with a fully connected layer followed by a softmax function to get a distribution over the output classes. The entire model is trained end-to-end using the cross entropy loss.

\section{Multi-task Learning}
\label{smtl}

\ignore{\yogarshicomment{This section conflates tasks and datasets and we should separate those. The first paragraph here should focus on the tasks, why you think those tasks will help you learn, and things like primary and auxiliary task can be defined here (primary = temporal rel. prediction, auxiliary = RE). In the next paragraph, we can discuss datasets in more gory detail.}}


While the model described in the previous section can be directly trained using labeled training data, the amount of annotated training data for this task (in the MATRES dataset) is limited. We enrich our model with useful information from other complementary tasks via SMTL. 

\subsection{Method}

We adapt the framework of \newcite{kiperwasser-ballesteros-2018-scheduled}, where three \textbf{schedulers} are used. They follow either a constant, sigmoid or exponential curve $p(t)$, where $p(t)$ is the probability of picking a batch from the main task, $t$ is the amount of data visited so far throughout the training process and $\alpha$ is a hyperparameter. The constant scheduler splits the batches randomly; at any time step, the model will be trained with sentences belonging to either the main task or the auxiliary task ($p^{const}(t) = \alpha$, $0 \leq \alpha \leq 1$) . The sigmoid scheduler allows the model to visit batches from both the auxiliary task and the main task at the beginning while the latest updates are always with batches consisting of batches from the main task ($p^{sig}(t) = \frac{1}{1+e^{-\alpha t}}$). The exponential scheduler starts by visiting only the batches from the auxiliary task while the latest updates are always from the main task ($p^{exp}(t) = 1 - e^{-\alpha t}$).

Following past work, we prepend a trained task vector to the encoder to help the model to differentiate between the main and the auxiliary tasks \cite[\textit{inter alia}]{ammar-etal-2016-many,johnson-etal-2017-googles,kiperwasser-ballesteros-2018-scheduled}.

\ignore{The six relations classes are 
\textit{Physical} (location of an entity or proximity of an entity with respect to another entity), 
\textit{Person-Social} (relationships between people),
\textit{Part-Whole} (when an entity is a part of another entity),
\textit{ORG-Affiliation} (when an entity is affiliated to or a member of another entity),
\textit{Agent-Artifact} (when an entity owns an artifact, has possession of an artifact, uses an artifact, or caused an artifact to come into being), and
\textit{Gen-Affiliation} (representing citizenship, residency, religion, or ethnic affiliation for people, or locations of an organization). The dataset has 
8,365 positive examples of the 6 annotated relation classes  and 79,147 negative examples of the class OTHER.}


\subsection{Auxiliary Datasets}

We use three different \textbf{auxiliary datasets} in our SMTL setup. The first two  have a different taxonomy and label set than MATRES, but have gold annotations. The last one is a silver dataset with predicted labels and same taxonomy as MATRES.

Our first dataset is the \textbf{ACE relation extraction task}.\footnote{\url{https://www.ldc.upenn.edu/sites/www.ldc.upenn.edu/files/english-relations-guidelines-v6.2.pdf}} We hypothesize that this task can add knowledge of different domains and of the concept of linking two spans in text given a taxonomy of relations. While this is not directly related to events and our farthest task in terms of similarity, the pairwise span classification is the reason we include this. 

We also use a closer and complementary temporal annotation dataset, i.e. the \textbf{Timebank and Aquaint annotations involving timex relations} (timex-event, event-timex, timex-timex)  \cite{ning-etal-2018-multi, goyal-durrett-2019-embedding}.\footnote{\url{http://www.timeml.org/publications/timeMLdocs/timeml_1.2.1.html}.} We expect the model to greatly benefit from being exposed to the timex relations in an MTL framework by learning about temporality in general and by adding specificity of the event-event temporal relations from the MATRES annotations. Figure \ref{timexexample} shows an example of the data annotated with an event-timex relation. 

\begin{figure}[t]
\fbox{
\begin{minipage}{7.4cm}
Robert F. Angelo, who \textbf{(event, left)} Phoenix at \textbf{(timex, the beginning of October)}.
\end{minipage}
}
\caption{Example of an event-timex annotation from the Timex annotations. The relation between \textbf{(event, left)} and \textbf{(timex, the beginning of October)} is \textit{Is$\_$included}.}
\label{timexexample}
\end{figure}



 
 We use self-training \cite{scudder1965probability} to generate our third dataset: \textbf{a silver dataset}. This requires an unlabeled text, a tagger to extract events from this text, and a classifier to predict temporal relations for pairs of extracted events. As our unlabeled text, we use 6,000 random documents from the CNN / Daily Mail dataset which is a collection of news articles collected between 2007 and 2015 \cite{hermann2015teaching}. We picked 85K segments of text within these documents that contain between 10 and 40 tokens after tokenization. We train a RoBERTa-based named entity tagger and use it to tag events in these segments.\footnote{The tagger is simply a dense layer on top of RoBERTa representation. We evaluate the tagger by using it to tag events in the MATRES validation set. The tagger reaches a F1 score of 89.5 on the MATRES development set.} This results in about 65K events. We consider all 285K pairs of events that lie within a segment as candidates for temporal ordering. Finally, we use our baseline RoBERTa temporal model to classify the temporal relation between these candidate pairs and use the top $2\over3$$^{rd}$ most confident classifications based on softmax scores to get about 190K instances of silver relations.

\section{Experiments and Results}
\label{results}


The MATRES dataset is our primary dataset for training and validation\ignore{\shuaicomment{for training (and validation)? This seems an incomplete sentence}}. As in previous work, we use TimeBank and AQUAINT (256 articles) for training, 25 articles of which are selected at random for validation and Platinum (20 articles) as a held-out test set \cite{ning-etal-2018-multi,goyal-durrett-2019-embedding,ning-etal-2019-improved}. Articles from TimeBank and AQUAINT at full length are about 400 tokens long on average. We believe that the document in its entirety is not required to infer the temporality between a given pair of events. Moreover, BERT style models are also often pre-trained for shorter inputs than this. For these reasons, we truncate our input text to a window of sentences\footnote{We use spacy \cite{spacy2} for sentence segmentation of the articles} starting with one sentence before the first event argument up to and including one sentence after the second event argument. 

We use one set of hyperparameters for all LSTM models and  another set for all the Transformer models (both with and without ELMo embedder).\footnote{LSTM models use 2 hidden layers with 256 hidden units each, and a batch size of 64. Transformer models use 1 hidden layer with 128 hidden units, and a batch size of 24. All models are trained using Adam \cite{Kingma2014AdamAM} with a learning rate of 10$^{-5}$ on an NVIDIA V100 16GB GPU.} BERT and/or RoBERTa are loaded as a replacement of the Transformer parameters and they are therefore used both as embedders and encoders. We run our SMTL and self training experiments with our best baseline model on the development data: the RoBERTa model.

For the SMTL experiments, we explore the $\alpha$  hyperparameter, and we pick the one that produces the highest scores in our development data. 

Finally, we picked our best SMTL model on the development data (see Table, this is the constant scheduler with silver data) parameters and continue training on the gold data only; we reduce the learning rate to 10$^{-6}$. This is because the model trained in the first step is already in a good state and we want to avoid distorting it with aggressive updates. 

We compare our results (Table \ref{restable}) with other top performing systems. First, we observe that among models without contextualized representations, the LSTM encoder is 2.5 F1 points better than the Transformer encoder. We observe that replacing static word representations with ELMo representations leads to significantly worse F1 with the LSTM encoder, but marginally improves upon the F1 of the Transformer encoder. We attribute this difference to the non-complementary nature of LSTM and ELMo representations, as ELMo is also LSTM-based, and thus the ELMo+LSTM combination might need more training data in order to extract meaningful signals.

Importantly, however, our base model that uses pretrained RoBERTa surpasses the previous state-of-the-art \cite{ning-etal-2019-improved} which uses BERT. Our BERT models yield very similar results to them. The main differences are that they do not finetune BERT along with the updates to the model, while we do and also, we model the context around the argument spans explicitly as part of $S_1$, $S_3$ and $S_5$. The reason why RoBERTa is better than BERT in this case is likely due to the fact that it has been trained longer, over more data, and over longer sequences. This matters because our temporal ordering model usually takes into account a long span in which both events occur.

\begin{table}[t]
\centering
\scalebox{0.75}{
\begin{tabular}{l|c|c}
\toprule
\textbf{Experiment} & \textbf{Acc} & \textbf{F1} \\
\midrule
LSTM  &  64.4 \footnotesize{$\pm$ 0.36} & 69.1 \footnotesize{$\pm$ 0.39} \\
+ Elmo &  60.0 \footnotesize{$\pm$ 2.89} & 64.8 \footnotesize{$\pm$ 3.00} \\
\hline
Transformer  &  61.9 \footnotesize{$\pm$ 0.93} & 66.4 \footnotesize{$\pm$ 0.99}\\
+ Elmo &  62.2 \footnotesize{$\pm$ 1.3} &  66.9 \footnotesize{$\pm$ 1.35}\\
\midrule
BERT base &  71.5 \footnotesize{$\pm$ 0.63} &  77.2 \footnotesize{$\pm$ 0.74} \\
\midrule
\midrule
RoBERTa base&  73.5 \footnotesize{$\pm$ 1.03}& 78.9 \footnotesize{$\pm$ 1.16}\\
+ SMTL (ACE) constant (0.6) &   72.5 \footnotesize{$\pm$ 0.69} & 78.5 \footnotesize{$\pm$ 0.84}\\ 
+ SMTL (ACE) exponent (0.5) &   71.5 \footnotesize{$\pm$ 1.81} & 77.4 \footnotesize{$\pm$ 1.19}\\
+ SMTL (ACE) sigmoid (0.5) &   70.0 \footnotesize{$\pm$ 1.81} & 76.4  \footnotesize{$\pm$ 0.89}\\
\hdashline
+ SMTL (Timex) constant (0.9) &   73.4 \footnotesize{$\pm$ 1.81} & 79.3  \footnotesize{$\pm$ 0.64}\\ 
+ SMTL (Timex) exponent (0.7) &   73.7 \footnotesize{$\pm$ 0.74} & 79.4  \footnotesize{$\pm$ -0.46}\\
+ SMTL (Timex) sigmoid (0.8) &   74.2 \footnotesize{$\pm$ 0.74} & 79.8  \footnotesize{$\pm$ 0.70}\\
\hdashline
+ SMTL (silver data) constant (0.05) & 73.8 \footnotesize{$\pm$ 0.74} & 80.3 \footnotesize{$\pm$ 0.51}\\
+ SMTL (silver data) sigmoid (0.2) &   74.0 \footnotesize{$\pm$ 0.73} & 80.1  \footnotesize{$\pm$ 0.72}\\
+ SMTL (silver data) exponent (0.1) &   73.9 \footnotesize{$\pm$ 0.64} & 79.6 
\footnotesize{$\pm$ 0.52}\\
\hdashline
Self-training: fine-tune on gold &  \textbf{75.5} \footnotesize{$\pm$ 0.39} & \textbf{81.6} \footnotesize{$\pm$ 0.26}\\
\midrule
\midrule
\newcite{ning-etal-2018-multi} & 61.6 & 66.6 \\
\newcite{goyal-durrett-2019-embedding}\footnote{\newcite{goyal-durrett-2019-embedding} report only \emph{accuracy} but they shared their confusion matrix so we calculate their F1 metric as in \newcite{ning-etal-2019-improved}.} & 68.6 & 74.2\\
\newcite{ning-etal-2019-improved} & 71.7 & 76.7 \\
\bottomrule
\end{tabular}
}
\caption{Results, including comparison with the best systems on the MATRES test set (Platinum). Results highlighted in bold are the best in each metric. We report average (and standard deviation) of accuracy and F1 over 5 runs with different random seeds. Given that it does not carry temporal information, we treat the relation VAGUE as a \emph{no relation} for the F1 results as in \newcite{ning-etal-2019-improved}. For the SMTL experiments, the selected $\alpha$ value is shown between parentheses.}
\label{restable}
\end{table}

The SMTL experiments show that the auxiliary task with timex annotations provides non-negligible improvements of almost 1 F1 point on top of our RoBERTa model. Learning from the timex annotations makes our model more aware of time relations and thus, better at ordering events in time. The sigmoid and exponent schedulers perform better than the constant scheduler, suggesting that the model needs to first learn about temporality, and then learn to be more specialized on predicting temporal ordering relations later. We believe this timex multi-tasking setup to be an implicit yet effective way to teach our model about timexes in general without timex embeddings used in \cite{goyal-durrett-2019-embedding}. When we use the ACE relation extraction dataset as an auxiliary task, none of the schedulers produce improvements while the sigmoid and exponent scheduler fare significantly worse. This result suggests that if the tasks differ too much, SMTL might not be a helpful strategy. 

The self-training experiments (including SMTL with silver data) show that the silver data helps to reach better performance with constant being the best scheduler. Furthermore, fine-tuning of the best model (according to development set score, which in this case it is the same as test set score) on the gold data gives us another boost in performance establishing a new state of the art in the task that is 2.7 F1 points better than our RoBERTa baseline, and almost 4 points better than the previous published results.


%
%




\section{Conclusions and Future Work}

This paper presents neural architectures for ordering events in time. It establishes a new state-of-the-art on the task through pretraining, leveraging complementary tasks through SMTL and self-training techniques. 

For the future, instead of using the RoBERTa baseline model for the self-training experiments, we could run several iterations by retraining on the data produced by our best self-trained model(s)\ignore{\miguelcomment{Nima: have you tried this?} \nimacomment{no, I felt the improvement of self-training over baseline is not enough that a second run would add something. In my past experiences (not for this paper) the noise would overcome the good signal in later iterations)}}; this could be a good avenue for further improvements. In addition we plan to extend our work by moving to other languages beyond English (we currently have not tried this due to lack of data) using cross-lingual models, \cite{subburathinam-etal-2019-cross}, applying other architectures like CNNs \cite{nguyen-grishman-2015-relation}, incorporating tree structure in our models \cite{miwa-bansal-2016-end} and/or by handling jointly performing event recognition and temporal ordering \cite{li-ji-2014-incremental,katiyar-cardie-2017-going}. 




\bibliography{acl2019,anthology}
\bibliographystyle{acl_natbib}

\end{document}